\documentclass[conference]{IEEEtran}
\IEEEoverridecommandlockouts
\usepackage{cite}
\usepackage{amsmath,amssymb,amsfonts}
\usepackage{algorithmic}
\usepackage{graphicx}
\usepackage{textcomp}
\usepackage{xcolor}
\usepackage{makecell}
\usepackage{pifont}
\usepackage{bbding}
\usepackage{fontawesome}
\usepackage{arydshln}
\usepackage{helvet}
\usepackage{siunitx}

\usepackage[a4paper, total={184mm,239mm}]{geometry}
\usepackage[linesnumbered,ruled]{algorithm2e}
\def\BibTeX{{\rm B\kern-.05em{\sc i\kern-.025em b}\kern-.08em
    T\kern-.1667em\lower.7ex\hbox{E}\kern-.125emX}}

\usepackage{tikz}
\newcommand*\circled[1]{\tikz[baseline=(char.base)]{
            \node[shape=circle,draw,inner sep=0.5pt] (char) {#1};}}

\begin{document}

\title{FineQ: Software-Hardware Co-Design for Low-Bit Fine-Grained Mixed-Precision Quantization of LLMs \\}

\author{
Xilong~Xie, Liang~Wang, Limin~Xiao, Meng~Han,  Lin~Sun, Shuai~Zheng, Xiangrong~Xu
}

\author{Xilong~Xie\textsuperscript{1}, Liang~Wang\textsuperscript{1}$^*$, Limin~Xiao\textsuperscript{1}$^*$, Meng~Han\textsuperscript{1}, Lin~Sun\textsuperscript{2}, Shuai~Zheng\textsuperscript{1}, Xiangrong~Xu\textsuperscript{1}\\
\textsuperscript{1} 
% State Key Laboratory of Software Development Environment \\
School of Computer Science and Engineering, Beihang University, Beijing 100191, China \\
\{xxl1399, lwang20, xiaolm, hanm, zhengshuai, xxr0930\}@buaa.edu.cn \\
\textsuperscript{2} Jiangsu Shuguang Optoelectric Co., Ltd \\ 
m13773555855@163.com

\IEEEcompsocitemizethanks{
\IEEEcompsocthanksitem $^*$ Liang Wang and Limin Xiao are corresponding authors of this paper.
\IEEEcompsocthanksitem This work was supported in part by the National Science and Technology Major Project under Grant No. 2022ZD0117602; in part by the National Natural Science Foundation of China under Grants 62272026 and 62104014; in part by the Fundamental Research Funds for the Central Universities; in part by State Key Laboratory of Complex \& Critical Software Environment under Grant No. CCSE-2024ZX-10.}
}

\maketitle

\begin{abstract}

Large language models (LLMs) have significantly advanced the natural language processing paradigm but impose substantial demands on memory and computational resources. Quantization is one of the most effective ways to reduce memory consumption of LLMs. However, advanced single-precision quantization methods experience significant accuracy degradation when quantizing to ultra-low bits. Existing mixed-precision quantization methods are quantized by groups with coarse granularity. Employing high precision for group data leads to substantial memory overhead, whereas low precision severely impacts model accuracy. To address this issue, we propose FineQ, software-hardware co-design for low-bit fine-grained mixed-precision quantization of LLMs. First, FineQ partitions the weights into finer-grained clusters and considers the distribution of outliers within these clusters, thus achieving a balance between model accuracy and memory overhead. Then, we propose an outlier protection mechanism within clusters that uses 3 bits to represent outliers and introduce an encoding scheme for index and data concatenation to enable aligned memory access. Finally, we introduce an accelerator utilizing temporal coding that effectively supports the quantization algorithm while simplifying the multipliers in the systolic array. FineQ achieves higher model accuracy compared to the SOTA mixed-precision quantization algorithm at a close average bit-width. Meanwhile, the accelerator achieves up to 1.79× energy efficiency and reduces the area of the systolic array by 61.2\%. 
\end{abstract}

% introduction 
% 

\section{Introduction}
Large language models (LLMs) based on transformers\cite{vaswani2017attention} have demonstrated remarkable performance in various natural language processing tasks through their enormous model size. However, LLMs, as one of the most significant computational workloads today, are challenging to deploy on edge devices due to their extensive memory and computational resources required\cite{luk2024hardware}. For instance, deploying the LLaMA-2-70B in FP16 requires at least 140 GB of memory\cite{touvron2023llama}, markedly exceeding the 80 GB capacity of an A100 GPU. Weight quantization \cite{frantar2022gptq,yao2022zeroquant,dettmers2022llm,lin2024awq,lee2024owq,li2023llm,shang2023pb,nagel2020up} is one of the most effective ways to reduce the memory consumption of LLMs while preserving performance, achieved by storing parameters in a low-precision representation.

% \vspace{-7mm}
\begin{figure}[htbp] 
\setlength{\abovecaptionskip}{-0.2cm}
\centering
% \vspace{-0.3cm}
{\includegraphics[width=0.48\textwidth]{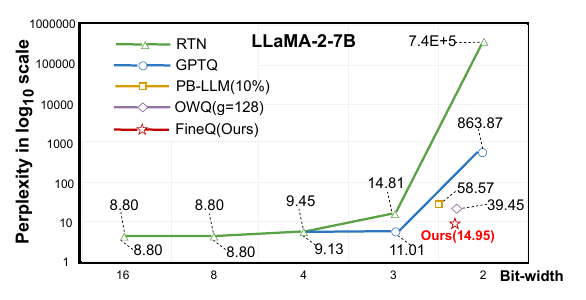}}
% \vspace{-0.2cm}
\caption{Perplexity of LLaMA-2-7B on C4 under differnet bit-widths. Lower perplexity means better model accuracy. Round-to-nearest (RTN), GPTQ, PB-LLM (10\% weight of FP16), and OWQ(g=128) suffer from accuracy loss at ultra-low bits. FineQ demonstrates better performance than these methods.}
\vspace{-0.5cm}
\label{Fig_quantization_error}
\end{figure}
% \vspace{-3.5mm}

To enable the future deployment of LLMs on edge devices\cite{laskaridismobile}, more aggressive quantization and compression techniques will be essential. Existing quantization methods are still inadequate for this purpose, as reducing precision to ultra-low bits($<$ 3 bits) leads to a significant degradation in model accuracy. Currently, weight quantization techniques primarily fall into single-precision quantization and mixed-precision quantization. Several single-precision quantization methods, such as RTN\cite{nagel2020up}, GPTQ\cite{frantar2022gptq}, have successfully quantized the FP16 LLMs into 4-bit or 3-bit formats while maintaining acceptable performance. Regrettably, as shown in Figure \ref{Fig_quantization_error}, advanced single-precision quantization methods for LLMs experience notable performance degradation under ultra-low bit quantization. This observation underscores the challenges that existing methods encounter when quantizing models to ultra-low bits.

The substantial loss of model accuracy in single-precision quantization is mainly due to the inadequate handling of weight outliers\cite{yao2022zeroquant}. Outliers are identified by a small set of values with exceptionally high magnitudes\cite{wei2023outlier}. Single-precision quantization methods indiscriminately clip both outliers and normal values, leading to notable drops in model accuracy\cite{dettmers2022llm}. To maintain model accuracy, recent works apply mixed-precision quantization methods. 
% \vspace{-7mm}
\begin{figure}[htbp] 
\setlength{\abovecaptionskip}{-0.2cm}
\centering
% \vspace{-0.4cm}
{\includegraphics[width = 0.44 \textwidth]{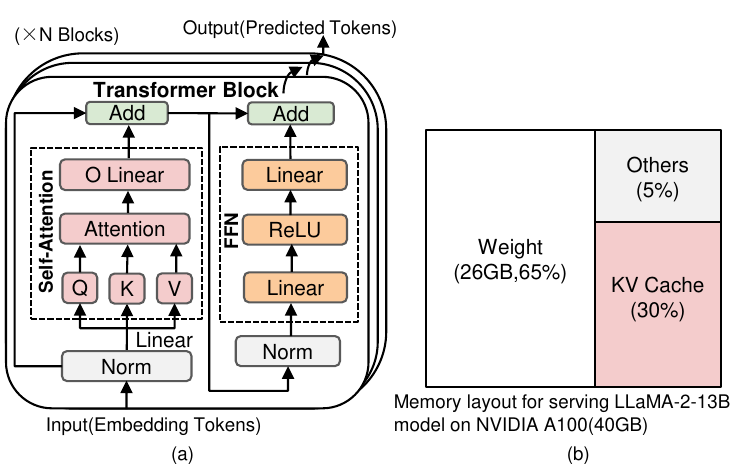}}
\caption{(a) The transformer block architecture. (b) Memory layout for serving a 13B-parameter LLM on the NVIDIA A100 (40GB)\cite{kwon2023efficient}.}
\vspace{-0.5cm}
\label{Fig_transformer}
\end{figure}
%\vspace{-2.5mm}
For instance, OWQ\cite{lee2024owq} and LLM-MQ\cite{li2023llm} quantize the majority
of normal weight values to 2 bits while representing outliers in FP16 format. PB-LLM \cite{shang2023pb} binarizes a subset of weights in LLMs while preserving outliers with high precision. Though existing mixed-precision quantization algorithms preserve model accuracy to some extent, they still demonstrate shortcomings in the handling of outliers. In existing mixed-precision quantization methods\cite{guan2024aptq,lee2024owq,shang2023pb,li2023llm}, weights are quantized in coarse-grained groups, with certain groups containing weights with large range. Therefore, these methods fail to adequately protect outliers and struggle to achieve a trade-off between memory overhead and model accuracy. Moreover, these methods treat outliers as sparse matrices, necessitating additional sparse formats and complex control logic to handle irregular computations and memory access, leading to lower computational efficiency.

In this work, we introduce FineQ, a low-bit fine-grained mixed-precision quantization method of LLMs with software-hardware co-design. We aim to achieve a balance between model accuracy and memory overhead by adopting a fine-grained outlier protection strategy. Specifically, we propose a series of innovative mechanisms and architectures, including a fine-grained mixed-precision quantization algorithm, an intra-cluster outlier protection mechanism, and a temporal coding-based hardware accelerator. The main contributions of this paper are summarized as follows.

1) We propose a fine-grained mixed-precision quantization algorithm that does not require retraining. The algorithm partitions model weights into smaller clusters and considers the distribution of outliers within each small cluster, enabling quantization to lower bit-width while maintaining model accuracy.

2) We introduce an intra-cluster outlier protection mechanism. By observing the distribution of weight outliers and the effects of various quantization bit-widths on model accuracy, we quantize the majority of normal values to 2 bits, apply 3-bit protection for outliers, and use an innovative memory alignment scheme that integrates encoding information with the data.

3) We present an efficient hardware accelerator based on temporal coding to support the quantization algorithm. Furthermore, temporal coding allows for efficient bitstream generation from weights\cite{wu2022ubrain}, significantly simplifying the multiplier design in the systolic array. The accelerator achieves up to 1.79× energy efficiency and reduces the systolic array area by 61.2\%.

\section{Background and Motivation}

\subsection{Transformer-Based Large Language Models}
% some introduction about LLMs
\begin{figure}[htbp] 
\setlength{\abovecaptionskip}{-0.2cm}
\centerline
{\includegraphics[scale=0.70]{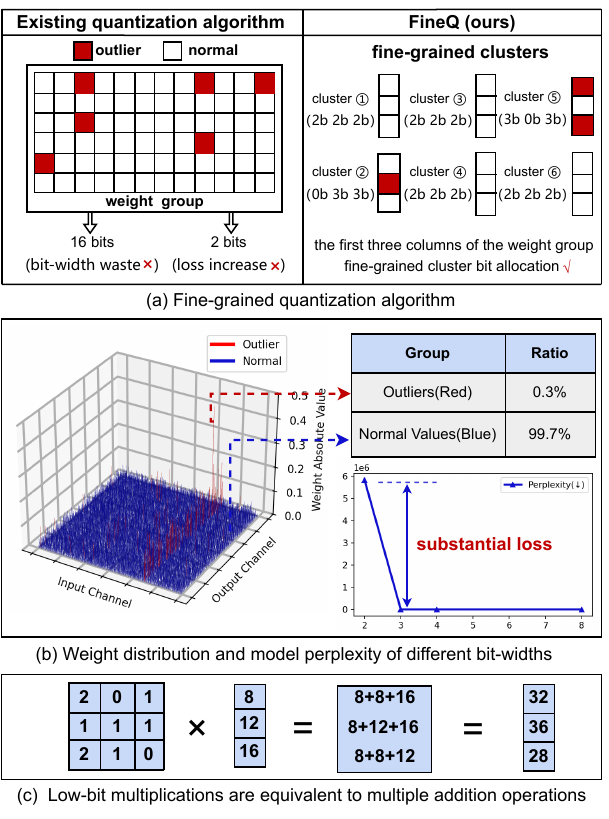}}
\caption{Three observations about hardware-software co-design of mixed-precision quantization.}
\vspace{-0.5cm}
\label{observation}
\end{figure}
The transformer model serves as the backbone architecture for existing LLMs\cite{vaswani2017attention}, enabling efficient processing for tasks such as natural language understanding and generation\cite{nassiri2023transformer,keswani2024abstractive}.
As illustrated in Figure \ref{Fig_transformer}(a), each transformer block contains self-attention and feed-forward network (FFN). The self-attention mechanism quantifies the degree of association among different positions within a single sequence and it consists of QKV generation, attention, and a linear layer. The FFN consists of two linear layers and an activation function.

In recent years, the scale of large language models has continued to grow\cite{zhao2023survey}. Figure \ref{Fig_transformer}(b) illustrates the memory distribution for a 13B-parameter LLM on an NVIDIA A100 GPU. Approximately 65\% of the memory is allocated for the model weights\cite{kwon2023efficient}. To facilitate the deployment of LLMs on memory-constrained edge devices, weight quantization emerges as one of the most effective approaches, as it can substantially reduce the memory overhead of large language models. However, weight outliers significantly impact model accuracy, making it challenging to quantize the model to ultra-low bits while maintaining model accuracy. Thus, effectively protecting weight outliers is crucial\cite{kim2023squeezellm,guan2024aptq}.

\subsection{Related Works}

Previous works\cite{frantar2022gptq,lin2024awq,lee2024owq,li2023llm,shang2023pb} on weight quantization, including single-precision and mixed-precision methods, aim to minimize memory overhead without affecting the accuracy of LLMs. Several single-precision quantization methods such as GPTQ\cite{frantar2022gptq} and AWQ\cite{lin2024awq} quantize the weight matrix in single-precision. Among them, GPTQ introduces an innovative one-shot weight quantization technique that leverages approximate second-order information. AWQ protects the salient weights by observing the distribution of activation values. Regrettably, these methods suffer from a significant precision loss when quantizing to ultra-low bits, due to outlier issues. In addition, some mixed-precision quantization methods are proposed to maintain model accuracy. For instance, OWQ\cite{lee2024owq} and LLM-MQ\cite{li2023llm} store weight outliers in FP16 Format separately from normal values. PB-LLM\cite{shang2023pb} binarizes a portion of the model's weights while preserving some outliers. However, these methods still suffer from accuracy loss due to inter-group data variability. Additionally, methods like Olive\cite{guo2023olive} and Tender\cite{lee2024tender} employ hardware-software co-design for outlier-aware quantization. They quantize both activations and weights to 4 bits and focus on outliers in activations rather than weights.

\subsection{Observation and Key Idea}
Through extensive exploration of existing implementations, we have three crucial observations that indicate significant optimization potential for hardware-software co-design of mixed-precision quantization, as shown in Figure \ref{observation}.

\textbf{Observation \uppercase\expandafter{\romannumeral1}: 
Mixed-precision quantization methods through coarse-grained groups struggle to strike a balance between model accuracy and memory overhead.} Existing mixed-precision quantization algorithms operate at the group level, with each group typically consisting of more than 128 weight values, and the distribution of outliers within the group is random. As illustrated in Figure \ref{observation}(a), a coarse-grained group may contain only a small number of outliers. Using a higher bit-width to store the entire group increases memory overhead, while a lower bit-width may fail to capture certain outliers, significantly impacting model accuracy.

\textbf{Key idea:} We propose a fine-grained intra-cluster quantization algorithm that splits weight data into smaller clusters, each containing only three weight values. Furthermore, we allocate bit-widths based on the data within each cluster, balancing both model accuracy and memory overhead.

\textbf{Observation \uppercase\expandafter{\romannumeral2}: Employing excessively high precision for weight outliers is unnecessary.} Figure \ref{observation}(b) visualizes the input weight distribution of a representative linear layer in the LLaMA-2-7B model\cite{touvron2023llama} and the model accuracy achieved with uniform quantization under different bit-widths\cite{dettmers2022gpt3}. The blue color represents normal values in the weights while the red color indicates outliers. Through observation, we found that over 99\% of the weights have very similar values. This indicates that only sparse outliers require additional processing, while the majority of weights can be quantized to lower bits. The distribution of outliers is not random but instead tends to concentrate within specific channels. Moreover, reducing the quantization bit-width from 3 bits to 2 bits significantly affects model accuracy. In contrast, increasing the bit-width from 3 bits to 16 bits has a limited impact on accuracy. 

\textbf{Key idea:} We apply quantization across weight channels and allocate 3 bits for outliers, further reducing the memory overhead without significantly impacting model accuracy.

\begin{figure}[htbp] 
\setlength{\abovecaptionskip}{-0.2cm}
\centerline
{\includegraphics[scale = 0.72]{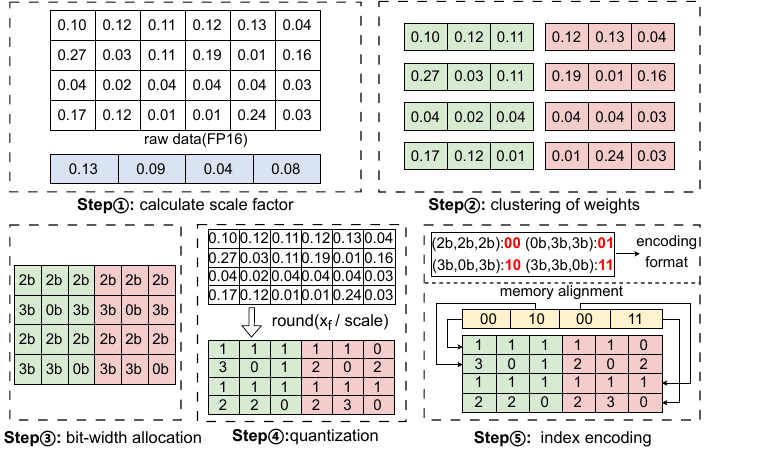}}
\caption{An example of fine-grained intra-cluster quantization algorithm.}
\vspace{-0.3cm}
\label{Fig_quantization_example}
\end{figure}

\textbf{Observation \uppercase\expandafter{\romannumeral3}: Low-bit quantization can significantly simplify the design of hardware multipliers.} Low-bit multiplication operations can be transformed into multiple accumulation operations. For instance, an $8\times2$ multiplication can be represented as an $8+8$ addition operation. Similarly, as illustrated in Figure \ref{observation}(c), low-bit matrix multiplication can be decomposed into multiple parallel accumulation operations without affecting the computational results. 

\textbf{Key idea:} We propose an efficient accelerator based on temporal coding, which encodes weight data into bitstreams. The accelerator significantly simplifies the multiplier design in the systolic array and improves hardware efficiency.

\section{Algorithm Design}
In this section, we introduce the fine-grained intra-cluster quantization algorithm, which partitions the weights into fine-grained clusters and separates them from other clusters to minimize quantization error. We propose an intra-cluster outlier protection mechanism that allocates 3 bits for outliers, effectively balancing memory overhead and model accuracy. Moreover, we provide a walking example of the algorithm.

\begin{algorithm}[]
\label{fine-grained_intra-cluster_weight_quantization}
\footnotesize
\caption{Algorithm for fine-grained intra-cluster weight quantization.}
\DontPrintSemicolon
  \SetAlgoLined
  \KwIn {Weight Matrix $W$; Layer Number $L$; } 
  \KwOut {Quantization Weight Matrix $Q$}
   \For {each layer $l$ and each channel $c$ in the weight     
  matrix $W_l$}{
    $W_{l, c} \gets$ weight vector of channel $c$ in layer $l$ \;
    $num\_clusters \gets \lceil \text{length}(W_{l, c}) / $3$ \rceil$ \;
    Divide $W_{l, c}$ into $num\_clusters$ of size $3$ \;
    \For {each cluster $C$ in $W_{l, c}$}{
        $max\_val \gets \max(C)$ \;
        $min\_val \gets \min(C)$ \;
        \If {$max\_val > 4 \times min\_val$}{
            Encode the top two values with $3$ bits \;
            Set remaining values to $0$ \;
        }
        \Else{
            Encoding all values in $C$ with $2$ bits \;
        }
        \ForEach{$C_i$ in clusters}{
        \eIf{$C_i$ has no neighbor}{
            Assign default encoding to $C_i$ \;
        }{
            \eIf{neighbor $C_j$ has encoding}{
                $E(C_i) \leftarrow E(C_j)$ \;
            }{
                $E(C_i), E(C_j) \leftarrow \arg\min_l Loss(C_i, C_j, l)$ \;
            }
        }
    }
        $Q \gets UniformQuantization(C)$ \;
    }
}

\end{algorithm}

\begin{figure}[htbp] 
\setlength{\abovecaptionskip}{-0.1cm}
\centering
{\includegraphics[width = 0.5 \textwidth, height=0.22\textheight]{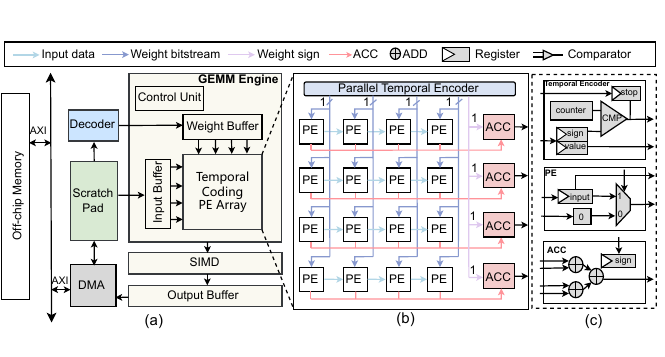}}
\caption{(a) Overview of FineQ architecture. (b) Temporal coding PE array design. (c) Temporal Encoder, PE, and ACC design.}
\label{Fig_architecture}
\vspace{-0.5cm}
\end{figure}

% \begin{figure*}[t]
% \setlength{\abovecaptionskip}{-0.2cm}
% \centerline{\includegraphics[scale=0.70]{figures/Fig_architecture.pdf}}
% \caption{(a) Overview of FineQ architecture. (b) Temporal coding PE array design. (c) Temporal Encoder, PE, and ACC design. }
% \label{Fig_architecture}
% \vspace{-0.5cm}
% \end{figure*}

\subsection{Fine-grained Intra-cluster Quantization Algorithm}

To balance memory overhead and model accuracy, we propose a fine-grained intra-cluster quantization algorithm, as illustrated in steps \circled{1} and \circled{2} of Figure \ref{Fig_quantization_example}. First, the scaling factor for each channel is computed according to Equation \ref{Uniform_quantization}, where $b$ is the bit-width, $s$ represents the scale factor, $x_{max}$ is the maximum value, $abs$ is the absolute value operation. Then, as shown in Figure \ref{observation}(b), most outliers are concentrated in specific channels. We perform per-channel quantization of weights and divide every three weights within each channel into a fine-grained weight cluster. The bit-width allocation for each cluster is determined exclusively by the data within that cluster. Before bit-width allocation within clusters, outliers within each cluster are identified by comparing the maximum and minimum values. If the maximum exceeds four times the minimum, the outlier protection mechanism is applied; otherwise, all cluster values are quantized to 2 bits. 

\begin{equation}
\setlength{\abovedisplayskip}{0.2cm}
\setlength{\belowdisplayskip}{0.2cm}
s=\frac{abs({x_{max}})}{2^{b-1}-1}; \quad
x_{q}=round\left(\frac{x_{f}}{s}\right)
\label{Uniform_quantization}
\end{equation}

\subsection{intra-cluster Outlier Protection Mechanism}

% bit allocation
The intra-cluster outlier protection mechanism is used to protect certain outliers in the model weights, thereby preserving model accuracy.
Figure 4 \circled{3} provides an example of bit-width allocation within a cluster. As shown in Figure \ref{observation}(b), quantizing to 3 bits strikes a balance between memory overhead and model accuracy. Unlike existing methods that typically use 16 bits to protect outliers\cite{lee2024owq,li2023llm}, we represent outliers with 3 bits. When outliers are detected within a cluster, the two largest values are encoded with 3 bits, while the smallest value is sacrificed.

Following bit-width allocation, as illustrated in Figure \ref{Fig_quantization_example} step \circled{4}, uniform quantization is performed according to Equation \ref{Uniform_quantization}, where $round$ is an approximation operation, $x_{f}$ and $x_{q}$ are a floating-point value and the quantized one. The bit-width allocation method effectively considers the distribution of outlier points within the cluster. The quantized data is converted from float to int, with outliers being effectively protected.

% encoding format
To distinguish the encoding scheme for each cluster, we use 2-bit encoding to specify the bit-width allocation: `00' indicates all values are 2 bits, `01' means the first value is zero with the remaining two encoded in 3 bits, `10' sets the second value to zero, and `11' sets the third value to zero. Additionally, we introduce a compression strategy that requires adjacent clusters to use the same encoding. If two adjacent clusters use different encoding schemes, we can apply fine-tuning to select the encoding. By iterating through four encoding schemes to select the one that minimizes the average error relative to the original data. As shown in Figure \ref{observation}(b), most weights are encoded in 2 bits, so fine-tuning is rarely needed. To ensure aligned memory access, we employ a specialized encoding scheme for both indices and data. As shown in Figure \ref{Fig_quantization_example} step \circled{5}. Specifically, four index values are encoded within a single byte to distinguish the encoding formats of the subsequent eight clusters. This approach reduces the overhead of storing sparse outlier indices compared to existing mixed-precision methods and facilitates more efficient memory access.  Moreover, the entire quantization algorithm is performed offline and the pseudocode is shown in Algorithm \ref{fine-grained_intra-cluster_weight_quantization}.

\section{Architecture Design}
% hardware accelerator
While the FineQ algorithm can be implemented in software, its full potential is realized through a custom accelerator design. In this section, we propose an efficient accelerator based on temporal coding. Our hardware design effectively supports fine-grained mixed-precision quantization algorithm through an efficient decoder unit that supports various cluster encoding schemes. Furthermore, we design a temporal coding PE array, which simplifies multipliers compared to conventional systolic array. Finally, we present a computational example.

\subsection{Architecture Overview}
Figure \ref{Fig_architecture}(a) presents the overview of FineQ architecture, it consists of a decoder unit, a vector processing unit, on-chip buffers for input/weight/output data, a direct memory access (DMA) unit, a temporal coding systolic array with input-stationary dataflow\cite{kwon2019understanding} and a control unit. The architecture features a six-stage pipeline: \textbf{(1) Off-chip Memory Access:} Weights and input data are read from off-chip memory and transferred to on-chip buffers via DMA. \textbf{(2) Decode:} The weight data is loaded into the buffer after passing through the decoder. \textbf{(3) Input Preloading:} The control unit sends control signals to the systolic array, while the input data is loaded into its registers. \textbf{(4) Matrix Multiplication:} Matrix multiplication is executed within the temporal coding systolic array, and the partial sums are then forwarded to the vector unit. \textbf{(5) Vector Processing:} Activation functions are executed by the vector unit. \textbf{(6) Write-back to Off-chip Memory:} Finally, the results are written back to the off-chip memory unit via DMA.

\subsection{Decoder Design}

\begin{figure}[htbp] 
\setlength{\abovecaptionskip}{-0.1cm}
\centering
{\includegraphics[width = 0.45 \textwidth]{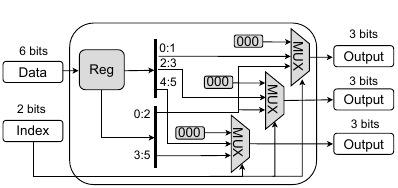}}
\caption{FineQ decoder design}
\label{Fig_decoder}
\vspace{-0.3cm}
\end{figure}

\begin{figure}[htbp] 
\setlength{\abovecaptionskip}{-0.1cm}
\centering
{\includegraphics[width = 0.48 \textwidth]{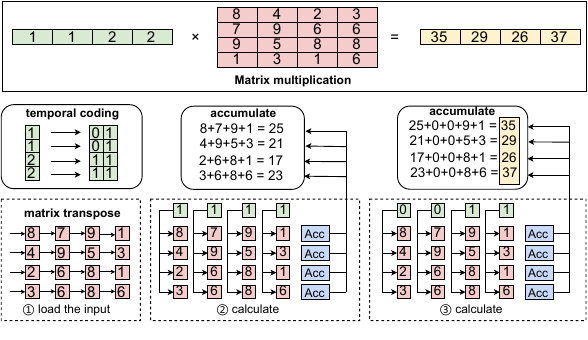}}
\caption{An example of matrix multiplication with temporal coding PE array.}
\label{Fig_compute_example}
\vspace{-0.5cm}
\end{figure}

The design of the weight decoder unit is shown in Figure \ref{Fig_decoder}. First, the input is divided into index and weight data, where the index is used to distinguish the quantization format of the cluster. Since the decoded data may be either 3 bits or 2 bits, we ensure all decoded data is 3 bits by padding with zeros. Next, the data is passed through multiple selection units. These units use the index information to output the decoded data and determine whether to perform zero-padding. Finally, the process decodes a cluster of data into three 3-bit weights.

\subsection{Temporal Coding PE Array Design}

Based on the idea that low-bit multiplication can be transformed into multiple accumulations, we propose a temporal coding PE array. It consists of parallel temporal encoder, processing elements (PEs), and accumulation units(ACCs). Temporal coding\cite{wu2022ubrain} is a lossless encoding method where the number of ones in the bitstreams represents the value. As shown in Figure \ref{Fig_compute_example}, the value 2 is encoded as '11' and the value 1 as '01'. Through temporal coding, low-bit data is converted into fixed-length bitstreams. 
Figure \ref{Fig_architecture}(b) shows the hardware architecture of the temporal encoder. The temporal encoder records the input data value, compares it with the counter, and outputs the result through a comparator. Additionally, the parallel temporal encoder can receive termination signals from the control module to stop bitstream generation. Then, the generated bitstreams are broadcast to each column of processing elements (PEs), and due to the 1-bit transmission width, the overhead associated with broadcasting remains minimal. Next, the PE stores the activation data. When weight bitstreams arrive, the selector determines whether to output the activation data or zero. In the accumulation unit, the sign bits of the weights and the outputs from each PE row are processed and accumulated using an adder tree structure. Overall, this design simplifies the MAC units of traditional systolic arrays and improves hardware efficiency.

Figure \ref{Fig_compute_example} also provides an example of matrix multiplication performed using temporal coding PE array. First, the control unit issues the control flow signal, and the input data is loaded into the PE units. Second, the bitstreams generated by the temporal encoder are broadcasted to the PE array. In step \circled{2}, the partial results are output by the accumulation unit, and in step \circled{3}, these results are reused and further accumulated until the computation is complete. It can be observed that the final computation results match those of the matrix multiplication.

\section{Evaluation}

In this section, we evaluate FineQ in terms of model accuracy, area, and energy efficiency.

\subsection{Evaluation Setup}
\textbf{Quantization Setup.} We choose LLaMA family(3B-13B)\cite{touvron2023llama} to evaluate our method. We use perplexity as the evaluation metric, which is a widely used one for LLMs. The perplexity of LLMs is evaluated on the WikiText2\cite{merity2016pointer} and C4 datasets\cite{raffel2020exploring}.  Lower perplexity means better model accuracy.

\textbf{Quantization Baseline.} 
We compare FineQ with five existing weight quantization methods: (1) Uniform \cite{dettmers2022gpt3} uses symmetric uniform quantization; (2) RTN\cite{nagel2020up} rounds all weights to the nearest quantized value on a fully uniform, asymmetric per-row grid; (3) 
GPTQ\cite{frantar2022gptq} leverages approximate second-order information; (4) PB-LLM\cite{shang2023pb} binarizes a subset of the weights; (5) OWQ\cite{lee2024owq} employs mixed precision including FP16 and INT2. RTN, Uniform, and GPTQ are single-precision methods, thus the bit width is set to 2. PB-LLM protects 10\% of the weight values, resulting in an average bit-width of 2.7 bits. OWQ employs a group size of 128, achieving an average bit-width of 2.25 bits. For FineQ, we reduce the weight bit-width to 2.33 bits. 

\textbf{Accelerator Baseline.} 
We compare the area and energy efficiency of the FineQ accelerator with the
systolic array using MAC Units. They have the same on-chip buffer size and use input-stationary dataflow\cite{kwon2019understanding}. The models used as workloads are from the quantization baseline.

\textbf{Architecture Implementation.}
We implement the temporal coding PE array and decoder using Verilog HDL and verify the design through RTL simulations. The temporal PE array consists of 4096 PEs and 64 Decoders. The accelerator is synthesized by Synopsys Design Compiler\cite{kurup1997logic} in a 45 nm process technology \cite{oliveira2016ascend}. We also develop a cycle-level simulator to estimate the overall performance.

\subsection{Accuracy Evaluation}

\textbf{Performance on LLMs.} We first evaluate the perplexity of the FineQ quantization algorithm, with an average bit-width quantized to 2.33 bits. As shown in Table \ref{Table_ppl_over}, FineQ outperforms existing methods on various models. Uniform, RTN, and GPTQ utilize single-precision quantization, and the results indicate that quantizing to 2 bits in these methods leads to significant loss due to the lack of protection from outliers. PB-LLM and OWQ employ mixed-precision quantization and are quantized by coarse-grained groups, making it difficult to effectively protect weight outliers under ultra-low bits. Furthermore, PB-LLM binarizes a substantial portion of the weights, employing a more aggressive strategy that results in a decrease in model accuracy. FineQ protects outliers through fine-grained quantization, thereby effectively maintaining model accuracy.

\begin{table*}[]
\renewcommand{\arraystretch}{1}
\centering
\caption{Perplexity results of quantized model on Wikitext2 and C4 datasets.}
\tiny
\label{Table_ppl_over}
\resizebox{\textwidth}{!}{
\begin{tabular}{c c| c c c c c c c}
\Xhline{0.2pt}
\hline
\multicolumn{2}{c}{\textbf{Model(PPL)}} && \multicolumn{2}{c}{\textbf{LLaMA-2-3B}}& \multicolumn{2}{c}{\textbf{LLaMA-2-7B}} &\multicolumn{2}{c}{\textbf{LLaMA-2-13B}}\\ \hline
Method &Avg. Bits &Seq. & Wiki & C4 & Wiki & C4 & Wiki & C4 \\ \hline
FP16  & {16}  & {2048}  & {7.35}  & {9.58} & {6.61} & {8.81} & {5.97} & {8.19} \\ 

RTN & {2} & {2048} & {1.6E+5} & {1.6E+5} & {4.3E+4} & {7.4E+5} & {6.3E+4} & {6.0E+4}\\ 

Uniform & {2} & {2048} & {6.3E+6} & {6.5E+6} & {5.8E+6} & {5.8E+6} & {2.6E+5} & {2.1E+5}\\ 

GPTQ &  {2}  &  {2048}  &  {1675.56} &  {5090.50}  &  {256.17} &  {863.87} &  {248.59}  &  {506.32}  \\ 

PB-LLM 10\% &  {2.7} & {2048} & {60.38} & {123.04} & {28.59} & {58.57}  & {131.54} & {208.34}\\ 

% group size = 128
OWQ(g128) &  {2.25} &  {2048} &  {34.51} &  {75.78} &  {22.95} & {39.45} & {15.19} & {26.03}\\

\textbf{FineQ(Ours)} &  {2.33} &  {2048} &  {\textbf{13.69}} & \textbf{{19.04}} &  \textbf{{10.94}} &  \textbf{{14.95}} &  \textbf{{13.16}} &  \textbf{{18.55}}\\ 
    \Xhline{0.2pt}
\hline

\end{tabular}
}
\vspace{-0.5cm}
\end{table*}

 \begin{table}[!htp]
    \centering
    \caption{Perplexity across different sequence lengths.}
    \LARGE
    \label{Table_ppl_len}
    \renewcommand{\arraystretch}{1.1}
    \resizebox{0.5\textwidth}{!}{
    \begin{tabular}{c c| c c c c c c}
    \Xhline{1.2pt}
    \hline
    \multicolumn{2}{c}{\textbf{SeqLen.(PPL)}} & \multicolumn{2}{c}{\textbf{32}}& \multicolumn{2}{c}{\textbf{256}} &\multicolumn{2}{c}{\textbf{1024}} \\ 
    \Xhline{1pt}
    \hline
    Method & Avg. Bits & Wiki & C4 & Wiki & C4 & Wiki & C4 \\ 
    \Xhline{1pt}
    \hline
    FP16  & {16} & {39.19} & {22.14} & {10.90} & {11.21} & {7.35}  & {9.19} \\ 
    
    RTN & {2} & {4.2E+4} & {3.5E+4} & {5.3E+4} & {5.5E+4} & {5.0E+4}  & {6.8E+4}\\ 
    
    Uniform & {2} & {4.3E+6} & {5.3E+6} & {5.0E+6} & {5.4E+6} & {5.4E+6}  & {5.3E+6}\\ 
    
    GPTQ &  {2} & {2.0E+5} & {1.7E+5} & {1.5E+5} & {1432.38} & {2.3E+5} & {1289.9} \\ 
    
    PB-LLM 10\% &  {2.7} & {286.13} & {271.18} & {52.60} & {73.19} & {32.41} & {58.97} \\ 
    
    OWQ(g128) &  {2.25} & {5.4E+4} & {6.3E+4} & {71.58} & {81.01} & {29.53} & {44.74} \\ 
    
    \textbf{FineQ(Ours)} & {2.33} & {\textbf{64.47}} & {\textbf{26.68}} & {\textbf{20.89}} & {\textbf{18.46}} & {\textbf{12.52}}  & {\textbf{15.77}}\\ 
    \Xhline{1.2pt}
    \hline
    \end{tabular}
    }
 \vspace{-0.5cm}
 \end{table}

\textbf{Sequence Length Sensitivity.} Table \ref{Table_ppl_len} presents a comparison of FineQ and previous methods on the LLaMA-2-7B model across three different sequence lengths. The results indicate that, while FineQ exhibits a slight increase in perplexity as sequence length decreases, it consistently outperforms prior methods in model accuracy. Additionally, FineQ exhibits greater robustness to variations in sequence length, demonstrating lower sensitivity to changes compared to other approaches.

\subsection{Performance and Area Evaluation}

\begin{figure}[htbp] 
\setlength{\abovecaptionskip}{-0.1cm}
\centering
{\includegraphics[width = 0.37 \textwidth]{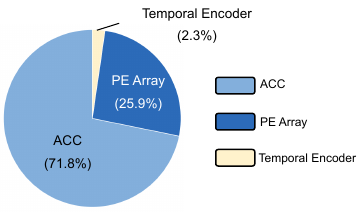}}
\caption{Breakdown of power in FineQ PE array.}
\label{Fig_power_percent}
\vspace{-0.3cm}
\end{figure}

\textbf{Area and Power.} We compare the accelerator area breakdown in Table \ref{Table_area_power}. The systolic array primarily consists of MAC units and accumulators. At the 400 MHz clock frequency and under the input-
stationary dataflow configuration, our temporal coding PE array reduces the area of the systolic array by 61.2\% and achieves a 62.9\% reduction in power consumption. Furthermore, as illustrated in Figure \ref{Fig_power_percent}, we present the power consumption distribution for each module in the FineQ PE array, where the Acc unit constitutes 71.8\%, the PE Array 25.9\%, and the Temporal Encoder accounts for only 2.3\%. The simplification of multipliers has significantly reduced the overhead of the PE array.

\vspace{-0.1cm}
 \begin{table}[!htp]
    \centering
    \caption{
    The area and power breakdown of accelerator core modules}
    \small
    \label{Table_area_power}
    \renewcommand{\arraystretch}{1.2}
    \resizebox{0.5\textwidth}{!}{
    \begin{tabular}{c|c|c|c}
    \Xhline{0.5pt}
        \hline
        Architecture & Setup &  Area  & Power\\
        \hline
        Systolic Array & 64×64 PEs& 0.954 $mm^2$ & 88.793 $mw$\\
        \hline
        FineQ Decoder & 64 &  0.008 $mm^2$&  0.187 $mw$\\
        \hline
        FineQ PE Array & 64×64 PEs &  0.370 $mm^2$ & 32.891 $mw$\\ 
        \hline
        \Xhline{0.5pt}
    \end{tabular}
    }
    \vspace{-0.15cm}
 \end{table}

\begin{figure}[htbp] 
\setlength{\abovecaptionskip}{-0.1cm}
\centering
{\includegraphics[width = 0.5 \textwidth]{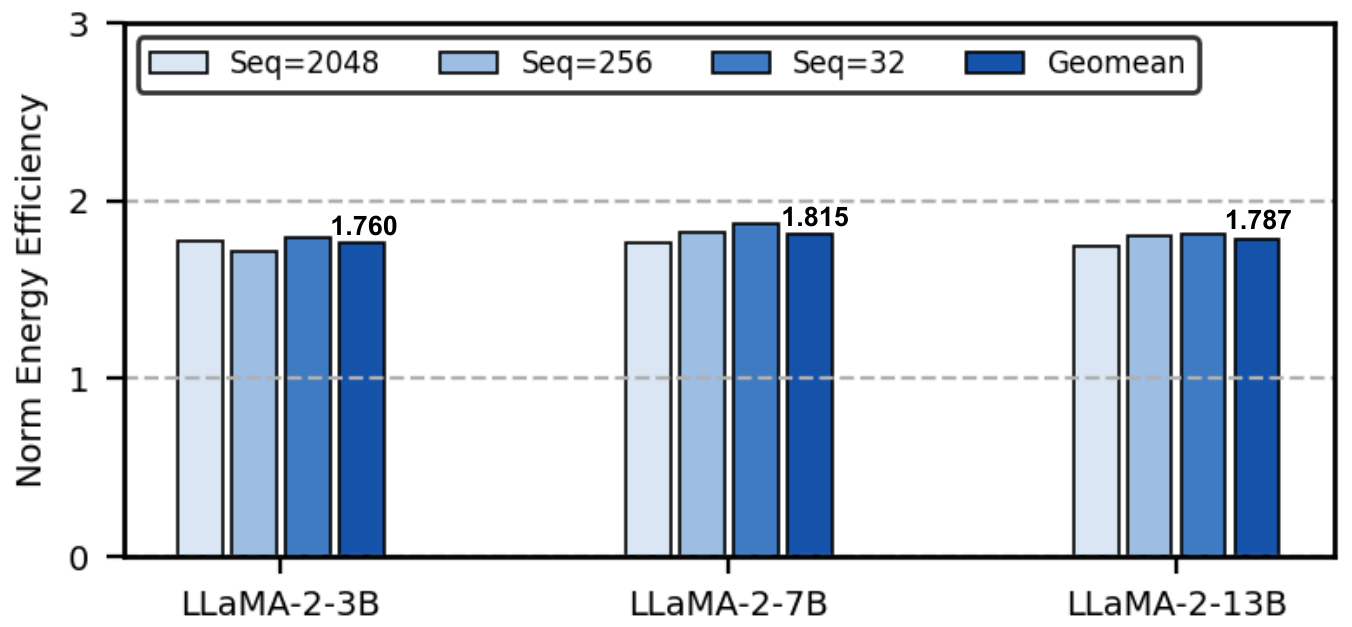}}
\caption{Normalized energy efficiency over baseline accelerator.}
\label{Fig_ef}
\vspace{-0.3cm}
\end{figure}

\textbf{Energy Efficiency.} Figure \ref{Fig_ef} illustrates the normalized energy efficiency of the accelerator at different sequence lengths under the same on-chip buffer size. Compared to the baseline design, the accelerator exhibits higher energy efficiency, attributed to the simplified design of most PEs and the reduction in the complexity of multipliers. Overall, the accelerator achieves up to 1.79× average energy efficiency on various models.

% \begin{figure}[htbp] 
% \setlength{\abovecaptionskip}{-0.1cm}
% \centering
% {\includegraphics[width=0.42 \textwidth]{figures/ef.pdf}}
% \caption{}
% \label{Energy_efficiency}
% % \vspace{-0.5cm}
% \end{figure}

\section{Conclusion} \label{Conclusion}

In this paper, we propose FineQ, a software-hardware co-design for low-bit fine-grained mixed-precision quantization of LLMs. The key insight is to handle outliers within clusters at a fine granularity. We address the issue of coarse quantization granularity in LLMs by proposing a fine-grained mixed-precision quantization algorithm, effectively balancing model accuracy and memory overhead. FineQ achieves higher model accuracy compared to the SOTA mixed-precision quantization algorithm at a close average bit-width. Additionally, we introduce an accelerator based on temporal coding to support this algorithm, simplifying multipliers in the systolic array and achieving up to 1.79× energy efficiency.

% \newpage
\bibliographystyle{IEEEtran}
\bibliography{mybib}

\end{document}